\documentclass[conference]{IEEEtran}
\IEEEoverridecommandlockouts
\usepackage{cite}
\usepackage{amsmath,amssymb,amsfonts}
\usepackage{algorithmic}
\usepackage{graphicx}
\usepackage{textcomp}
\usepackage{xcolor}
\usepackage[colorinlistoftodos]{todonotes}
\def\BibTeX{{\rm B\kern-.05em{\sc i\kern-.025em b}\kern-.08em
    T\kern-.1667em\lower.7ex\hbox{E}\kern-.125emX}}
\begin{document}

\title{Sensorless Hand Guidance using Microsoft Hololens
}

\author{
\IEEEauthorblockN{David Puljiz,  Erik St\"ohr, Katharina S. Riesterer, Bj\"orn Hein, Torsten Kr\"oger }
\IEEEauthorblockA{
\textit{Karlsruhe Institute of Technology}\\
Karlsruhe, Germany \\
david.puljiz@kit.edu, ujeji@student.kit.edu, uxecu@student.kit.edu, bjoern.hein@kit.edu, torsten@kit.edu}
}

\maketitle

\begin{abstract}

Hand guidance of robots has proven to be a useful tool both for programming trajectories and in kinesthetic teaching. However hand guidance is usually relegated to robots possessing joint-torque sensors (JTS). Here we propose to extend hand guidance to robots lacking those sensors through the use of an Augmented Reality (AR) device, namely Microsoft's Hololens. Augmented reality devices have been envisioned as a helpful addition to ease both robot programming and increase situational awareness of humans working in close proximity to robots. We reference the robot by using a registration algorithm to match a robot model to the spatial mesh. The in-built hand tracking capabilities are then used to calculate the position of the hands relative to the robot. By decomposing the hand movements into orthogonal rotations we achieve a completely sensorless hand guidance without any need to build a dynamic model of the robot itself. We did the first tests our approach on a commonly used industrial manipulator, the KUKA KR-5.  

\end{abstract}


\section{Introduction}
Hand guidance has become ubiquitous in collaborative robots such as the Universal Robotics' UR series or KUKA's LBR iiwa, mostly due to the safety requirements needing sensors, usually joint-torque sensors, on each joint and the ease of programming such a modality allows. The easy teaching of new trajectories is a great boon for robots in flexible manufacturing. Implementation of hand guidance in industrial robots in the mean time has been marginal, usually requiring external force-torque (FTS) or other sensors to accomplish the task. Thus the traditional teach pendant is used in the vast majority of cases for programming industrial robots, which is much more time consuming. Besides industrial settings, hand guiding is useful in imitation learning \cite{argall2009survey}, which can greatly reduce the search space in learning new tasks, as it already receives positive examples from demonstration.  \par

The use of AR in the field of robotics is a long but sporadic research field, with most of the research using either camera and screen or a projector to display information. This is due to the fact that head-mounted devices (HMDs) weren't robust or practical enough to be used. The emergence of practical HMDs has allowed more flexible applications not bounded to an already set-up environment. Wearable AR has been used to ease robot programming, knowledge patching for imitation learning, task planning for collaborative human-robot workspaces and for the display robot information to humans, among other things. \par

Other methods of implementing sensorless hand guidance already exist. Moe et al. use a Microsoft Kinect and a smartphone-based accelerometer to perform hand tracking and guide the end effector of an industrial robot \cite{Moe2013hg}. This approach however was limited to 5DOF. Furthermore by just driving the end effector one can not make use of the extra degrees of freedom a robot with seven or more joints provide. Lee et al. proposed a method for a completely sensorless system where industrial manipulators can be guided in a similar fashion \cite{Lee2016sensorless}. However the approach requires experiments to determine the friction model of each robot. Furthermore the robot is confined to using a torque controller. Finally the external force needed to move the end effector was found to be 1.23-4.83 times greater than approaches based on joint-torque sensors. Ideally the force would be close to zero, which isn't the case even for JTS methods. \par 

Here we present our research on an easily transportable method, not requiring any external sensors or sensors on the robot. The method requires only a kinematic model of the robot. No experiments before the use or any adaptation to different robot models are required. Furthermore it works with any controller, the external force needed to move the robot is zero as no force needs to be directly applied to the robot, and the sensitivity of the movement can be swiftly modified on-line. Finally it can be easily coupled with other wearable-AR-based robot programming and visualization modalities. \par

\section{Methodology}

The system consists of three main parts - the Hololens, a PC and the KR-5 robot. The computer is running the open-source Robot Operating System (ROS), and is connected to the KR5-arc. The workflow is as follows - the AR device is started and connects to the computer running ROS, which sends it the Universal robot description file (urdf) and associated visual and collision meshes. This allows the device to generate a holographic robot with the same geometry and kinematics as the real one. \par

The user then chooses either to place the robot manually or to use semi-automatic referencing. The manual mode likely won't result in a good match between the real robot and the hologram, and is intended more in the cases where the robot is inaccessible or when it's hard to manually reach particular links due to the size of the robot. In the latter case where the robot is too large it can also be resized. \par

The semi-automatic mode requires the user to place a "seed" hologram in the robot base and orient it to the front of the robot. We assume that the robot is static at the moment this is done, which is known to the controlling computer and sent to the AR device as well. This seed hologram is then used as both the center of the mesh to be extracted from the Hololens' spatial mapping mesh and the first guess for the registration algorithm. Once the mesh is sampled into a point-cloud it is send to the ROS computer where it's processed using the Point Cloud Library (PCL) by removing outliers and resampling it with the Moving Least Squares method (MLS). Then it is registered either via Iterative closest point (ICP) or via Super4PCS \cite{super4pcs}. The final transformation is then published via the tf node and accessible to both the ROS nodes running on the computer and the Hololens itself. This allows for referencing between the real robot and the Hololens, that is the real robot and the user's hands. \par

We make convex meshes around each joint to define the "collision" zone where a user's hand movements influence the joint positions of the robot. The actual movement of the connected joint is calculated based on the vector of the user’s hand positions from the previous to the current frame. Because of this, no movement is possible in the first frame after a hand enters a collision zone. A plane is defined using the affected joint’s position and the normal which is the joint's rotation axis. The vector of the hand movement is projected onto this plane and the projected vector is used to calculate the joint movement. If the joint movement doesn't violate any joint limits, as defined in the urdf, it is sent to the robot for execution, otherwise it's kept as is. In the vast majority of cases however, the entire hand movement vector cannot be performed by a single joint. In that case a new point is calculated by rotating the hand position in the previous frame around the joint by the calculated joint angle update. A new vector is that calculated between this point and the hand position in the current frame. This vector is then passed down the kinematic chain and the process repeated until we've performed the desired movement or we've run out of joints. This method allows us to handle an arbitrary number of joints. This is useful in the case of seven or more joints, as the extra degrees of freedom can be used to move around obstacles etc.

The vector of joint position updates is then sent to ROS where any control can be selected through ROS Control. In our work we primarily used the Reflexxes interpolated joint position controller, as it provided the most robust performance.

\section{Experiments}

Basic experiments were conducted on the KR-5. The registration proved quite robust as can be seen in table \ref{table:1}. The quality metric was the RMS of the distances between the closest point in the matched robot and the scene pointcloud. The big signifies a pointcloud created from a spatial mesh with 1,240,000 triangles per cubic meter and 256,000 samples per mesh; the small one is 1000 and 16.000 respectively. Registration allows us to operate the real robot just as though it had JTSs. The Super4PCS, being a global registration algorithm and therefore needing a segmentation of the input pointcloud, performed poorly in robot working cells where the robot was very close to another object e.g. a table, as the other object couldn't be segmented out. The robot performed the desired movements and a close match could be seen between the hand movement and the movement of the robot. A small user study was also conducted. Here however, the lack of end effector dragging meant that the tasks were completed slowly. Additionally the robot itself didn't have any extra DoF to benefit from our approach. Thus end effector dragging is one of the main additions to this method in the future.

\begin{table}{
\caption{Statistics of the conducted parameter tests of ICP, Super4PCS in combination with MLS. The statistics are based on RMS of the distances between the closes points in the matched robot and the original scene pointcloud }
\label{table:1}
\begin{tabular}{ |l|l|l|l|l|l|l|l| } 
 \hline
 Algorithm&Min&Max& Mean&Standard Deviation\\ 
 \hline
 ICP-big & 0.0012 & 0.4560 & 0.0039 & 0.01774 \\ 
 ICP-small & 0.00162 & 0.01441 & 0.0032 & 0.0017 \\ 
 Super4PCS-big & 0.0004 & 0.0947 & 0.0174 & 0.02907\\
 Super4PCS-small & 0.0002 & 0.0674 & 0.01034 & 0.00777\\
 \hline
\end{tabular}
}

\end{table}

\section{Conclusion and Future Work}

Up till now, we used the inbuilt Hololens' hand tracking. This has a few drawbacks, namely the user needs to use predefined gestures which may be unnatural. Secondly the hand tracking has issues when the hand is in close proximity of the robot, as well as having problems near dark surfaces. Losing hand tracking can have big impact on the stability of the control. We plan to switch to the CNN presented in \cite{GANeratedHands_CVPR2018} in the future. Implementation of additional features, primarily end effector dragging, and the ability to scale hand movements online, is also planed. The stability of the referencing and the scalability of the approach should be tested with other robot models in addition to the KR-5. Finally more extensive user studies to see how well our method compares to similar ones should be conducted. 


\section*{ACKNOWLEDGMENT}
This work has been supported from the European Union’s Horizon 2020 research and innovation programme under grant agreement No 688117 “Safe human-robot interaction in logistic applications for highly flexible warehouses (SafeLog)”.



\bibliographystyle{IEEEtran}
\bibliography{bibliography}

\end{document}